\documentclass{article}

\usepackage{microtype}
\usepackage{booktabs} 

\usepackage{hyperref}

\usepackage[accepted]{icml2021}

\icmltitlerunning{Controllable Graph Generation}

\usepackage{amsmath,amsthm,amssymb, multirow, array, bm} 
\usepackage[capitalize]{cleveref}
\usepackage[export]{adjustbox}
\usepackage{graphicx, subcaption}
\usepackage{stfloats}

\usepackage{empheq} 

\begin{document}

\twocolumn[
\icmltitle{\textsc{ShadowCast}: Controllable Graph Generation}



\icmlsetsymbol{equal}{*}

\begin{icmlauthorlist}
\icmlauthor{Wesley Joon-Wie Tann}{soc}
\icmlauthor{Ee-Chien Chang}{soc}
\icmlauthor{Bryan Hooi}{soc}
\end{icmlauthorlist}

\icmlaffiliation{soc}{Department of Computer Science, National University of Singapore}

\icmlcorrespondingauthor{Wesley Tann}{wesleyjtann@u.nus.edu}

\icmlkeywords{Controllable Graph Generation, Machine Learning, ICML}

\vskip 0.3in
]



\printAffiliationsAndNotice{}  

\begin{abstract}
We introduce the controllable graph generation problem, formulated as controlling graph attributes during the generative process to produce desired graphs with understandable structures. Using a transparent and straightforward Markov model to guide this generative process, practitioners can shape and understand the generated graphs. 
We propose \textsc{ShadowCast}, a generative model capable of controlling graph generation while retaining the original graph's intrinsic properties. The proposed model is based on a conditional generative adversarial network. Given an observed graph and some user-specified Markov model parameters, \textsc{ShadowCast} controls the conditions to generate desired graphs.
Comprehensive experiments on three real-world network datasets demonstrate our model's competitive performance in the graph generation task. Furthermore, we show its effective controllability by directing \textsc{ShadowCast} to generate hypothetical scenarios with different graph structures. 

\end{abstract}

\section{Introduction}
In many real-world networks, including but not limited to communication, financial, and social networks, graph generative models are applied to model relationships among actors. It is crucial that the models not only mimic the structure of observed networks but also generate graphs with desired properties because it allows for an increased understanding of these relationships. Currently, there are no such methods designed for the effective control of graph generation.

Meaningful interactions between agents are often investigated under different what-if scenarios, which determines the feasibility of the interactions under abnormal and unforeseen circumstances. In such investigations, instead of using actual data, we can generate synthetic data to study and test the systems~\citep{10.5555/956415.956464, 6890935}. However, there are many challenges. $(1)$ Data is not accessible by direct measurement of the system. $(2)$ Data is not available. $(3)$ Data produced by generative models cannot be understood. To address these challenges, we have to answer a natural and meaningful question: Can we control the generative process to shape and understand the generated graphs?

\begin{figure*}[tbp]
        \centering
        \includegraphics[width=.8\linewidth]{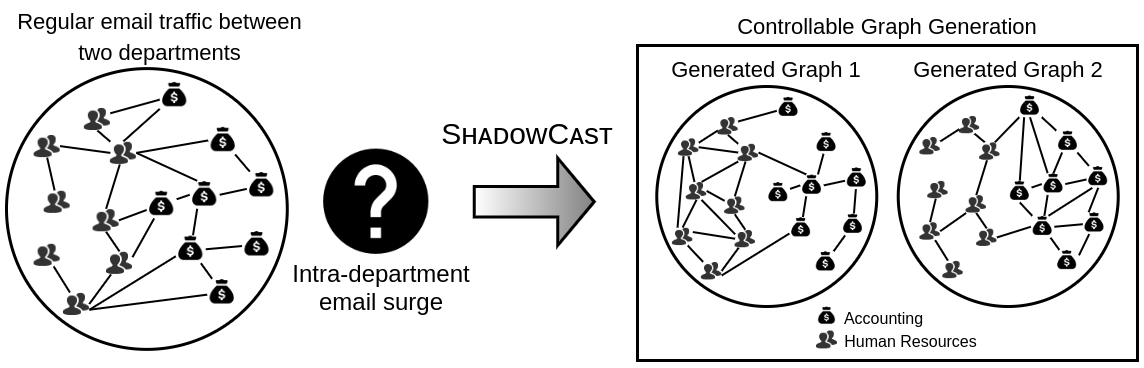}
        \caption{\textbf{Case study illustration of controlling graph generation}: Many times, data of various situations are not available in observed real-world networks. \textsc{ShadowCast} allows us to generate graphs of desired structures and provide an understanding of the generated graphs.}
        \label{fig:casestudy}
    \end{figure*}

In this work, we introduce the novel problem of controlling graph generation. The goal is to generate graphs of desired shapes by learning to control the associated graph attributes and structure to influence the generative process. We provide an illustrative case study of email communications in an organization with two departments (Figure~\ref{fig:casestudy}), where interactions of the employees follow a regular pattern during normal operations. Due to limited data, previously observed network information may be missing scenarios of intra-department email surge within either the Human Resources or Accounting departments. When such situations are required for analyzing the system, an ideal model should generate graphs that reflect these scenarios (see box in Figure~\ref{fig:casestudy}) while maintaining the organizational structure. By effectively controlling the generative process, \textsc{ShadowCast} allows users to generate designed graphs that meet conditions resembling a wide range of possibilities. 
Overall, this is a meaningful problem because controlling the generative process to understand generated networks proves to be valuable in many applications such as anomaly detection and data augmentation.

Existing graph generative models aim to mimic the structure of given networks, but they cannot easily shape graphs into other desired states. These works either directly capture the graph structure~\citep{cao2018molgan,liu2017gan, Tavakoli2017LearningSG,10.3389/fdata.2019.00003,NIPS2018_7942,You2018GraphRNNGR,Simonovsky2018GraphVAETG,pmlr-v80-bojchevski18a} or model node feature information~\citep{kipf2016variational,Wang2018GraphGAN,Grover2019GraphiteIG,Zou_2019}. Most of them adopt implicit model approaches, such as the popular generative adversarial networks (GANs)~\citep{goodfellow2016nips}. Only very recent advances~\citep{li2018learning,NIPS2019_8415} in network generation have started injecting auxiliary information into the model by adding graph-level conditions as additional inputs. However, none of them allow direct control over the generative process, which addresses the fundamental challenge of controlling generated graphs.


Another series of work on graph translation attempts to learn a translation mapping from the input domain to the target graph domain.
These methods either perform graph topology translation or predict the node attributes using the learned translation mappings. On the one hand, node attributes translation prediction~\citep{NIPS2016_3147da8a,li2018diffusion,ijcai2018-0505,10.1145/3282866.3282872,jin2018learning} aims to predict node attributes given a graph with fixed topology. On the other hand, graph topology translation~\citep{1805-09980,pmlr-v89-sun19c,8970898} studies the changing of graph topology domain distributions, forming target graph topologies by assuming that the node attributes are fixed. 
These works in graph topology translation are closest to our problem, where models generate graphs by discovering the underlying translation rules between input and target graphs. Even though translation mappings can create graphs, they only replicate real target graphs and cannot produce graphs with different structures without a provided target graph.

We propose \textsc{ShadowCast}, an approach for controlling graph generation, which addresses the challenge of generating graphs with user-desired structures. It is achieved by using easy-to-understand node attributes that are intended to capture graph semantics in an explicable way. These attributes form the shadow that we control in order to guide the graph generative process. 
The model architecture is essentially based on conditional GANs~\citep{DBLP:journals/corr/MirzaO14}. The model introduces control by leveraging the conditions, which we manage with a transparent Markov model, as a control vector to influence the generative process. It allows for user-specified parameters such as density distributions to generate designed graphs that are understandable. Finally, the generator captures essential graph structures while exploring a myriad of other possibilities in multifarious networks.  

We first evaluate \textsc{ShadowCast} on three real-world social and information networks to demonstrate its competitive performance against several state-of-the-art graph generation methods in mimicking given graphs. 
Our model achieves impressive results that are superior in most datasets. In addition, we demonstrate the capability of \textsc{ShadowCast} to produce customized synthetic graphs with different structures through tunable and intuitive parameters, which existing generative models are not designed to perform effectively.


\section{Controllable Graph Generation}
\label{gen_inst}
In this section, we describe the controllable graph generation problem. 
The core idea of the problem lies in generating graphs of desired structures through the control of its node attributes. 
We define these attributes and the graph structure as a \textit{shadow} and introduce our approach \textsc{ShadowCast}. 
Since it is a challenge to directly control the generation of graphs due to their complex interconnected nature, we model them through shadows, which can be manipulated to control the graph generation. 
We depict the problem and our approach in detail below (\cref{formulate,proposed}).

\subsection{Problem Formulation}
\label{formulate}
We focus on the novel problem of controlling graph generation. Let $\mathcal{G} = (\mathcal{V}, \mathcal{E})$ denote a graph with $N$ nodes $v_i \in \mathcal{V}$ and $E$ edges $(v_i, v_j) \in \mathcal{E}$. 
Each node is associated with some identity information, e.g., the employee ID. 
In addition, we induce another graph with $N$ nodes and the same edge connections as in $\mathcal{G}$, which we define as \textit{shadow} $\mathcal{S}$. 
Each of the nodes in $\mathcal{S}$ is associated with some property label $k_i \in K$, e.g., the employee's department, and it ``shadows'' the corresponding node in $\mathcal{G}$. Every node in $\mathcal{G}$ can be uniquely identified by the identity, whereas the label of each node in $\mathcal{S}$ is not necessarily unique. 
Using a transparent and straightforward Markov model to direct this shadow, we control the generative process and shape graphs with understandable outcomes while preserving the original graph properties. 
We note that there could be other properties of interest, e.g., degree distribution, a shadow with different connectivity than $\mathcal{G}$. We leave the inclusion of additional properties as extensions for future work. 

In this work, we aim to develop a controllable network graph generative model. By training the model $\bm{\Theta}$ on a graph $\mathcal{G}$ and its shadow $\mathcal{S}$, the model would then monitor the generative process and subject the generation to direction---aiding in the controllability of the generated graphs. 
Let us define the \textit{Controllable Graph Generation} problem as such: 
\begin{center}
\noindent\parbox{0.8\linewidth}{%
    Given a graph $\mathcal{G}$ and its key node attributes $K$, induce another graph with the same structure, defined as \textit{shadow} $\mathcal{S}$, where nodes in $\mathcal{S}$ are labeled by $K$. Train model $\bm{\Theta}$ to learn a representative shadow $\tilde{\mathcal{S}}$, and control $\tilde{\mathcal{S}}$ to generate graphs $\tilde{\mathcal{G}}$'s with understandable structures.
}%
\end{center} 

Following this process, we can leverage node properties such as ground-truth labels and other node attributes, valuable in understanding the model-generated results, as a control vector to guide the graph generation.

\subsection{Proposed Model}
\label{proposed}

    \begin{figure*}[htbp]
        \centering
        \includegraphics[width=.8\linewidth]{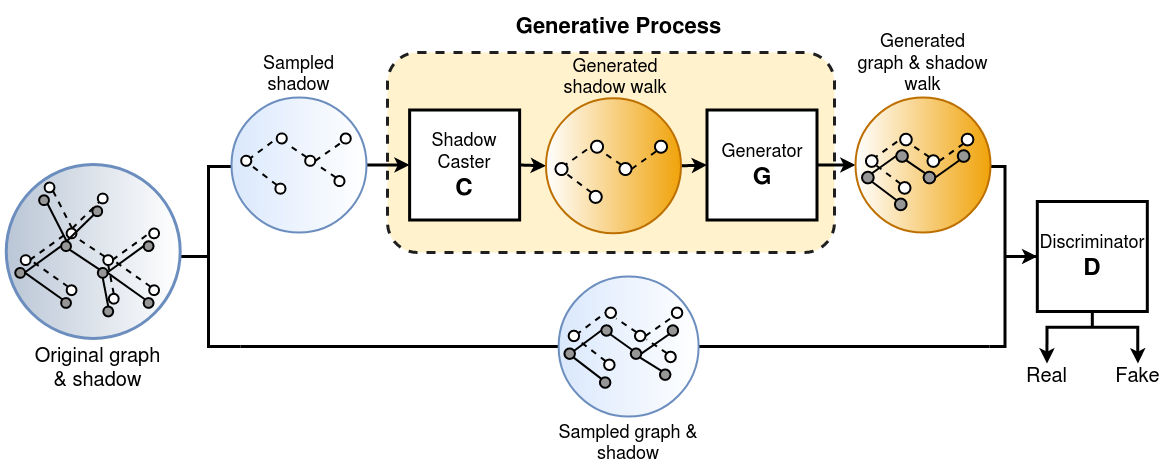}
        \caption{The \textsc{ShadowCast} architecture proposed in this paper.}
        \label{fig:ShadowCast_architecture}
    \end{figure*}

We propose \textsc{ShadowCast}, a controllable generative model that leverages both conditional modeling and GANs to generate graph-structured data. 
Our approach is inspired by the graph generative model~\citep{pmlr-v80-bojchevski18a} that poses the graph generation problem as learning a distribution of biased \textit{random walks} over the input graph. However, \citet{pmlr-v80-bojchevski18a} is not a controllable generative method.

In contrast, we design \textsc{ShadowCast} as a controllable graph generation method. It is based on a conditional GAN framework, which leverages the node attributes as a control vector to influence the generative process.
During the training phase, \textsc{ShadowCast} takes an attributed graph $\mathcal{G}$ as input and induces a shadow $\mathcal{S}$, where nodes in $\mathcal{S}$ correspond to attributes $K$. 
Next, the model trains shadow caster $C$, which takes in sampled sequences $\bm{s}$ of $\mathcal{S}$ and produces representative shadow walks $\tilde{\bm{s}}$ that mimic the real shadow.
Generative model $G$ then takes these representative shadow walks as conditions and generates fake samples of graph walks and shadow walks. 
The goal of $G$ is to capture the distribution of walks over the graph and generate synthetic graph random walks and conditions that are similar to the real walks. 
At the same time, discriminative model $D$ estimates the probability that a graph random walk and its corresponding conditions came from the real graph rather than $G$, to distinguish between the synthetic and real walks. 

After training the model, we can specify Markov model parameters ($\bm{\pi}, \bm{A}$) for controlling shadow caster $C$, producing representative shadow walks following the specified dynamics to create a representative shadow $\tilde{\mathcal{S}}$.
We then direct the generative process with this designed $\tilde{\mathcal{S}}$ and influence generator $G$ to generate desired graphs $\tilde{\mathcal{G}}$'s. 
We provide details of our model architecture (Figure~\ref{fig:ShadowCast_architecture}) and design choices below.


Using the conditional GAN framework, we train both $G$ and $D$ conditioned on some extra information---sampled shadow walks $\bm{s}$ from the shadow $\mathcal{S}$. By allowing our model to consider any auxiliary information such as ground-truth communities or data from other sources, the model can $(1)$ leverage extra information from different data modalities, and $(2)$ directly control the data generation process. For example, by using contextual information in the social communications of an organization, we learn semantically meaningful graph representations. We can then explicitly generate networks of any given context.

Following~\citet{DBLP:journals/corr/MirzaO14}, we introduce the conditional GAN training for graph shadow random walks and define the loss as:
\begin{equation} \label{eq:cganloss}
    \mathcal{L}_{cgan} = \log (D(\bm{x} \mid \bm{s})) + \log (1 - D(G(\bm{z} \mid \bm{s})))
\end{equation}   

where $\bm{z} \sim \ \mathcal{N}(\bm{0}, \bm{I}_d)$ is a latent noise from a multivariate standard normal distribution. We represent a social transaction network as an input graph of $N$ nodes as a binary adjacency matrix $\bm{\bar{A}} \in \{0,1\}^{N \times N}$. We then sample sets of random walks of length $T$ from $\bm{\bar{A}}$ to use as training data $\bm{x}$ for our model. Following~\citet{pmlr-v80-bojchevski18a}, we use a biased second-order random walk sampling strategy~\citep{10.1145/2939672.2939754}---one of the advantageous properties of random walks is their invariance under node reordering---in order to better capture both global and local graph structures. Another advantage of random walks is that the walks only include connected nodes, which efficiently exploits the sparsity of real-world graphs by including nonzero values of the adjacency matrix $\bm{\bar{A}}$. In the rest of this section, we describe in detail each stage of the \textsc{ShadowCast} generation process and formally present the procedure (\cref{alg:xggan}).

    \begin{figure}[htbp]
    \centering
    \begin{minipage}{\linewidth}
    \begin{algorithm}[H]
    \caption{Minibatch stochastic gradient descent training of controllable graph generative adversarial nets. The number of steps to apply to the generator, $\omega$, is a hyperparameter. We used $\omega = 3$.
    }\label{alg:xggan}
    \begin{algorithmic}
        \FOR{number of training iterations} 
        \STATE Sample minibatch of $m$ samples $\{\bm{x}^{(1)}, \dots, \bm{x}^{(m)}\}$ from data distribution $p_{data}$
        \STATE Sample the respective $m$ shadow walks $\{\bm{s}^{(1)}, \dots, \bm{s}^{(m)}\}$ 
        \STATE Update $C$ model weights: 
        \STATE  \begin{center}$\nabla_{\theta_s} \frac{1}{m} \sum\limits_{i=1}^{m} [- \sum\limits_{c=1}^{K} \bm{s}^{(c)} \log C(\bm{s}^{(c)})]$ \end{center}
        
        \STATE Generate minibatch of $m$ shadow walks $\{\tilde{\bm{s}}^{(1)}, \dots, \tilde{\bm{s}}^{(m)}\}$ with model $C$ 
        \STATE Sample minibatch of $m$ noise samples $\{\bm{\bm{z}}^{(1)}, \dots, \bm{\bm{z}}^{(m)}\}$ from $\mathcal{N}(\bm{0}, \bm{I}_d)$
        \STATE Update $G$ model weights: 
        \STATE  \begin{center}$\nabla_{\theta_g} \frac{1}{m} \sum\limits_{i=1}^{m} [\log (1 - D(G(\bm{\bm{z}}^{(i)} \mid \tilde{\bm{s}}^{(i)})))]$ \end{center}
        \FOR{$\omega$ steps} 
        \STATE Update $D$ model weights: 
        \STATE \begin{center} $\nabla_{\theta_d} \frac{1}{m} \sum\limits_{i=1}^{m} [\log D (\bm{x}^{(i)} \mid \tilde{\bm{s}}^{(i)}) + \log (1 - D(G(\bm{\bm{z}}^{(i)} \mid \tilde{\bm{s}}^{(i)})))]$  \end{center} 
        \ENDFOR  %
        \ENDFOR   %
    \end{algorithmic}
    \end{algorithm}  
    \end{minipage}
    \end{figure}

\paragraph{Shadow Caster} 
The shadow caster $C$ is a sequence-to-sequence model that learns arrays of contiguous node properties from sampled shadow walks on the shadow. The network predicts a sequence of inputs one at a time when some sequence is observed. We model $C$ with a long short-term memory (LSTM)~\citep{10.1162/neco.1997.9.8.1735} neural network. Given sampled sequences of shadow walks $(\bm{s}_1, \hdots, \bm{s}_{T})$ from the shadow as inputs, the shadow caster $C$ then generates synthetic shadow walks $(\tilde{\bm{s}}_1, \hdots, \tilde{\bm{s}}_{T})$ to mimic the sampled walks.

\paragraph{Generator}
The generator $G$ is a probabilistic sequential learning model that generates conditional graph random walks $(\bm{v}_1, \hdots, \bm{v}_T) \sim G$. We model $G$ using another parameterized LSTM network $f_\theta$. At each step $t$, $f_\theta$ takes as input the previous memory state $\bm{m}_{t-1}$ of the LSTM model, the current additional information $\tilde{\bm{s}}_t$, and the last node $\bm{v}_{t-1}$. The model produces two values $(\bm{p}_t, \bm{m}_t)$, where $\bm{p}_t$ denotes the probability distribution over the current node and $\bm{m}_t$ the current memory state. 
Next, the current node $\bm{v}_t$ is sampled from a categorical distribution $\bm{v}_t \sim Cat(\sigma(\bm{p}_t))$ using a one-hot vector representation, where $\sigma(\cdot)$ is the softmax function.

In order to initialize the model, we draw a latent noise from a multivariate standard normal distribution $\bm{\bm{z}} \sim \ \mathcal{N}(\bm{0}, \bm{I}_d)$ and pass it through a hyperbolic tangent function $g_{\theta'}(\bm{\bm{z}})$ to compute memory state $\bm{m}_0$. Generator $G$ takes as inputs the noise $\bm{\bm{z}}$ and sampled shadow walks $\bm{s}$, and it outputs graph random walks $(\bm{v}_1, \hdots, \bm{v}_T)$. Through this process, $G$ generates fake random walks.


\paragraph{Discriminator}
The discriminator $D$ is a binary classification LSTM model. The goal of $D$ is to discriminate between real walks sampled from walking on the original graph and fake walks generated by $G$. At each time-step $t$, the discriminator takes two inputs: the current node $\bm{v}_t$ and the associated shadow $\bm{s}_t$, both represented as one-hot vectors. After processing each presented sequence of shadow and graph walks, $D$ outputs a score between $0$ and $1$, indicating the probability of a real walk. 

After training the model, we have a shadow caster $C$ and a generator $G$ that can produce synthetic graphs. The shadow caster first constructs shadow walks $(\tilde{\bm{s}}_1, \hdots, \tilde{\bm{s}}_{T})$ of some user-defined class distribution (a relatively small number of shadow walks, e.g., 10,000). The generator then takes $(\tilde{\bm{s}}_1, \hdots, \tilde{\bm{s}}_{T})$ and generates a large set of random graph walks (a much larger number of random walks than for training, e.g., 10M). We construct a score matrix $\bm{S}$ by counting how often an edge appears in the set of graph walks. Next, we convert $\bm{S}$ into a binary adjacency matrix $\bm{\hat{A}}$ by first setting $s_{ij} = s_{ji} = \max \{s_{ij} , s_{ji}\}$ to get a symmetric matrix. Next, we could use simple binarization strategies such as thresholding or choosing top-$k$ entries. However, we follow a probabilistic strategy, introduced in~\citet{pmlr-v80-bojchevski18a}, that mitigates the issue of leaving out the low-degree nodes and producing singletons because the starting nodes of every walk is random.

\subsection{Controlling Generated Graphs} 
Different from existing approaches, our model takes shadow walks---a series of random walks on the node properties graph---as inputs to the generator, and it creates graphs with various densities. To answer questions like: ``Why did the model generate such graphs? Could we modify it to our desire?'', we generate graphs that are more understandable by controlling these shadow walk inputs. Our goal is to provide controllability and gain insight into how black-box generative models produce graphs.  

For any desired graph, we first build a Markov chain to model and construct sequences of node properties
based on some user-specified transition distribution. These sequences are then injected into the shadow caster $C$ to generate shadow walks $(\tilde{\bm{s}}_1, \hdots, \tilde{\bm{s}}_{T})$ that mimic the original shadow. Next, given a trained \textsc{ShadowCast} model $\bm{\Theta}$ and the shadow walks $(\tilde{\bm{s}}_1, \hdots, \tilde{\bm{s}}_{T})$, the generator $G$ produces desired graphs $\tilde{\mathcal{G}}$'s. Through this process, one can control the shadow distributions and study the generated graphs with varying structures.


\section{Related Work} 
\label{related}
Although many existing works study the generalizability of graph generation methods, effectively controlling graph generation remains an open question. From a broader perspective, we can consider the related problems of (1) constructing generative models for graph-structured data and (2) learning translation mappings between input and target graphs to infer their results.

\paragraph{Graph Generation}
Most existing graph generation models are designed to generate graphs mimicking the structure of observed graphs. So far, no generative method that shapes graphs into new desired states have been proposed. In general, we can group these graph generative models into two main families---those that directly model the graph structure~\citep{cao2018molgan,liu2017gan, Tavakoli2017LearningSG,10.3389/fdata.2019.00003,NIPS2018_7942,Simonovsky2018GraphVAETG} and others that study the graph in the context of node representations~\citep{kipf2016variational,Wang2018GraphGAN,Grover2019GraphiteIG,Zou_2019}. While modeling of graph structures approximates the distribution of graphs with minimal assumptions about their structure, modeling node embedding estimates the probabilities of each edge’s existence, which effectively models the relational structure of large graphs. 


Recently, some works in graph generation have started exploring network structures of various conditions. These works employ graph-level condition information. In one work, \citet{li2018learning} produce some conditional generation results, where the conditions are graph properties such as the number of nodes and edges. Another work, CondGEN~\citep{NIPS2019_8415}, injects semantics into the graphs by conditioning the model on supplementary contextual information. The model mainly considers multiple small graphs, each with an accompanying semantic condition to learn a distribution over graphs. While GraphRNN~\citep{You2018GraphRNNGR} is not a direct conditional model, it decomposes the generative process into sequences of nodes, which potentially allows for explicit conditioning. However, these methods only generate graphs mimicking the observed graphs.

To allow state manipulation and controllable graph generation, our model borrows the concept from NetGAN~\citep{pmlr-v80-bojchevski18a}, which adapts the standard LSTM to learn a distribution of random walks and exploit sparsity in real-world graphs. In contrast to NetGAN, we integrate a condition-based control mechanism to learn a model that generates custom graphs. Due to the challenging nature of the problem, to the best of our knowledge, no work has definitively considered shaping graphs into new desired states.



\paragraph{Graph Translation}
Another series of tangential work, graph translation, attempts to learn a translation mapping from the input domain to the target graph domain. These methods, using translation mappings, either perform graph topology translation or predict the node attributes. Existing graph translation models~\citep{NIPS2016_3147da8a,li2018diffusion,ijcai2018-0505,10.1145/3282866.3282872,jin2018learning} learn to predict node attributes given a graph with fixed topology. While predicting node attribute values is useful to model the behavior of nodes, potentially leading to enhanced performance in areas such as anomaly detection, it is limited in generating graphs.

Topology translation methods~\citep{1805-09980,pmlr-v89-sun19c}, which study the task of changing graph topology domain distributions, form target graph topology (i.e., structure, edges) by assuming that the node attributes are fixed. \citet{1805-09980} introduces a GAN-based graph translator using graph convolution, deconvolution layers, and a conditional discriminator to learn the global and local translation mapping. \citet{pmlr-v89-sun19c} proposes a conditioned graph generation model with two edge generation options. It is based on the GraphRNN model that learns the attention on input graph annotations, sequentially generating the nodes to form a target graph.

However, these models either predict node or edge attributes given fixed topology and fixed node attributes, respectively. They cannot simultaneously perform both predictions. \citet{8970898} develops an end-to-end framework for such joint prediction to circumvent the limitations of existing graph translation models, integrating both node and edge translations.


\section{Experiments}
\label{exp}
In this section, we first compare and evaluate our approach with other baseline graph generation methods on three datasets to establish our model's ability to generate high-quality graphs of complex networks. Next, we demonstrate the controllability of \textsc{ShadowCast} by directing the generative process to create graphs according to specifications. Note that generating graphs mimicking any given graph as closely as possible is not our goal. Our objective is to introduce a controllable graph generative approach that provides insights into generated graphs. Through our experiments, we not only demonstrate that \textsc{ShadowCast} exhibits competitive performance in the task of graph generation, but we also show that our model can generate graphs of different density distributions by controlling the shadows.

    \begin{table*}[htp]
    \centering  
    \caption{Performance statistics (mean and standard error) of the graphs generated by \textsc{ShadowCast} and the baseline models, computed over five runs. We indicate the mean values of the generated statistics closest to the real graphs. \textsc{ShadowCast} most closely matches original graphs in the statistics when compared with the baseline models.
    \label{table:stats}}    
    \resizebox{\linewidth}{!}{%
    \begin{tabular}{c|c|cccccc}
        \textbf{Graph}  & \textbf{Model} & 
        \multicolumn{1}{c}{\textbf{\begin{tabular}[c]{@{}c@{}}ASST\end{tabular}}} &
        \multicolumn{1}{c}{\textbf{\begin{tabular}[c]{@{}c@{}}CLUST\end{tabular}}} &
        \multicolumn{1}{c}{\textbf{\begin{tabular}[c]{@{}c@{}}CPL\end{tabular}}} &
        \multicolumn{1}{c}{\textbf{\begin{tabular}[c]{@{}c@{}}GINI\end{tabular}}} &
        \multicolumn{1}{c}{\textbf{\begin{tabular}[c]{@{}c@{}}MD\end{tabular}}} &
        \multicolumn{1}{c}{\textbf{\begin{tabular}[c]{@{}c@{}}TC\end{tabular}}} 
        \\ \hline
        \multirow{6}{*}{\textit{Cora-ML}}   & Real  &\textit{-0.075} &\textit{0.00277} &\textit{5.636} &\textit{0.485} &\textit{241.0} &\textit{2898.0} \\ \cline{2-8}
            & GraphRNN         &0.062$\pm$5.5e-4 &0.00121$\pm$2.2e-7 &1.892$\pm$5.7e-5 &0.119$\pm$1.9e-4 &507.4$\pm$2.7 &9023.8$\pm$17.8  \\
            & GVAE         &-0.324$\pm$6.1e-3 &0.01294$\pm$4.2e-4 &3.481$\pm$1.1e-2 &0.825$\pm$1.1e-3 &121.6$\pm$7.0 &15513.0$\pm$186.1  \\
            & NetGAN  &-0.055$\pm$1.5e-3 &0.00140$\pm$2.8e-5 &4.943$\pm$9.5e-3 &0.407$\pm$1.3e-3 &223.6$\pm$2.1  &1034.6$\pm$18.7  \\
            & CondGEN         &-0.524$\pm$2.0e-2 &0.00524$\pm$9.0e-4 &2.168$\pm$1.7e-2 &0.946$\pm$1.5e-3 &404.0$\pm$37.0 &95843.4$\pm$4780.4  \\
            & \textsc{ShadowCast}    &\textbf{-0.081$\pm$3.1e-3} &\textbf{0.00191$\pm$1.5e-4} &\textbf{5.187$\pm$1.0e-2} &\textbf{0.459$\pm$1.3e-3} &\textbf{229.6$\pm$7.7} &\textbf{1713.6$\pm$26.4}  \\ \hline 
        
        \multirow{6}{*}{\textit{Enron}}  & Real     &\textit{-0.003} &\textit{0.03300} &\textit{2.154} &\textit{0.281} &\textit{74.0} &\textit{4784.0} \\ \cline{2-8}
            & GraphRNN         &0.028$\pm$6.3e-3 &0.02154$\pm$3.1e-4 &1.977$\pm$5.5e-3 &0.116$\pm$2.0e-3 &30.8$\pm$0.8 &1221.6$\pm$34.4  \\
            & GVAE    &-0.112$\pm$2.1e-2 &0.04625$\pm$1.0e-3 &\textbf{2.165$\pm$6.9e-3} &0.288$\pm$7.2e-3 &45.2$\pm$1.3 &5439.2$\pm$58.7  \\
            & NetGAN  &0.123$\pm$1.2e-2 &0.03051$\pm$3.4e-4 &2.105$\pm$3.8e-3 &0.244$\pm$5.8e-3 &55.8$\pm$1.4 &3486.0$\pm$57.9  \\
            & CondGEN         &-0.287$\pm$2.7e-2 &0.04074$\pm$1.5e-3 &2.102$\pm$2.3e-2 &0.463$\pm$7.1e-3 &70.4$\pm$1.7 &9619.6$\pm$183.7  \\
            & \textsc{ShadowCast}    &\textbf{-0.004$\pm$4.6e-3} &\textbf{0.03483$\pm$7.8e-4} &2.214$\pm$6.2e-3 &\textbf{0.278$\pm$1.8e-3} &\textbf{73.2$\pm$2.6} &\textbf{5262.2$\pm$42.8}  \\ \hline
    
        \multirow{6}{*}{\textit{EUcore-top}}& Real  &\textit{-0.085} &\textit{0.03105} &\textit{2.885} &\textit{0.433} &\textit{65.0} &\textit{8133.0} \\ \cline{2-8}
            & GraphRNN         &-0.005$\pm$8.6e-3 &0.00891$\pm$1.1e-4 &2.128$\pm$5.2e-3 &0.118$\pm$9.6e-4 &41.0$\pm$0.84 &2255.2$\pm$59.2  \\
            & GVAE    &-0.257$\pm$1.2e-2 &\textbf{0.02919$\pm$3.8e-4} &2.579$\pm$7.0e-3 &0.473$\pm$2.4e-3 &68.8$\pm$2.3 &9025.2$\pm$127  \\
            & NetGAN  &-0.028$\pm$1.0e-2 &0.02335$\pm$3.1e-4 &2.642$\pm$1.0e-2 &0.359$\pm$1.7e-3 &62.0$\pm$2.2 &4639.8$\pm$28.8  \\
            & CondGEN         &-0.378$\pm$3.8e-2 &0.01880$\pm$1.9e-3 &2.101$\pm$1.2e-2 &0.720$\pm$3.6e-3 &147.0$\pm$10.0 &26106.4$\pm$726.4  \\
            & \textsc{ShadowCast}    &\textbf{-0.034$\pm$1.1e-2} &0.02847$\pm$3.4e-4 &\textbf{2.843$\pm$1.0e-2} &\textbf{0.435$\pm$2.6e-3} &\textbf{66.2$\pm$1.1} &\textbf{7414.4$\pm$93.8}    
    \end{tabular}   }
    \end{table*}

\paragraph{Datasets}\label{data}
We consider three real-world graphs in social and information networks, where each node belongs to one of the ground-truth communities. Two of the datasets are email communication networks \textit{EUcore-top} ($N$ = 348, $E$ = 3342, $\lvert K \rvert$ = 5) and \textit{Enron} ($N$ = 154, $E$ = 1843, $\lvert K \rvert$ = 3). The other dataset \textit{Cora-ML} ($N$ = 2810, $E$ = 7981, $\lvert K \rvert$ = 7) is a commonly used subset of a large author citation dataset. We provide the links to datasets used in our experiments (see Appendix for details).

We study communication networks: (1) \textit{EUcore-top} is a network that consists of the top five largest departments in the EUcore email dataset that was created using anonymized emails from a large European research institution. 
(2) \textit{Enron} is a dataset of the Enron email corpus where nodes are employees labeled according to their department information. 
The citation network: (3) \textit{Cora-ML} is a popular benchmark citation dataset. Nodes labeled according to their paper topic are authors, and edges between them indicate that an author cited another author's paper.

\paragraph{Baselines} 
Since guiding the generative process to provide controllable graph generation is a novel task, and no such method is developed, we compare our approach against four current state-of-the-art graph generation baseline methods---GraphRNN~\citep{You2018GraphRNNGR}, GVAE~\citep{Simonovsky2018GraphVAETG}, NetGAN~\citep{pmlr-v80-bojchevski18a}, and CondGEN~\citep{NIPS2019_8415}. We randomly select $85\%$ of the edges in each graph for training and use the remaining $15\%$ for validation and testing. 
We refer readers to the Appendix for more details about the model implementation settings, baseline models, datasets, and generated graph visualizations.

\paragraph{Performance}
We evaluate \textsc{ShadowCast} against existing benchmark generative models~\citep{You2018GraphRNNGR, Simonovsky2018GraphVAETG, pmlr-v80-bojchevski18a, NIPS2019_8415} and present the comparison statistics~\footnote{Statistics measuring graphs generated by \textsc{ShadowCast} and the baseline methods include ASST (assortativity), CLUST (clustering coefficient), CPL (character path length), GINI (Gini index), MD (maximum node degree), and TC (triangle count).} (\cref{table:stats}). By comparing the statistics of the real graphs and those generated by each method, closer mean values indicate greater resemblance to the original graphs, thus better performance. In general, baseline methods succeed at replicating the graphs that are directly modeled. Unsurprisingly, GVAE, designed for generating small graphs, performs well in the smaller \textit{Enron} and \textit{EUcore-top} datasets. However, it does not recover statistics of the larger graph \textit{Cora-ML} well. On the other hand, our model captures all graph properties of the datasets, especially excelling in preserving properties of larger graphs, as shown in its generation of the \textit{Cora-ML} dataset.

\textsc{ShadowCast}, a conditional generative model that considers meaningful auxiliary information (e.g., node labels) of given graphs on top of learning the graph structure, naturally outperforms methods that take an unconditional approach. The baseline methods are designed to generate graphs unconditionally, with the exception of CondGEN. However, CondGEN performs conditional generation with graph-level conditions, which are not as informative as the node-level information we inject into \textsc{ShadowCast}. This rich supplementary node information enables our model to learn better representations of graphs. Hence, \textsc{ShadowCast} achieves such outstanding performance.

    \begin{figure*}[htp] 
    \centering 
    \begin{adjustbox}{minipage=\linewidth,scale=0.8}
      \begin{subfigure}[b]{0.5\linewidth}
        \centering
        \includegraphics[width=\linewidth]{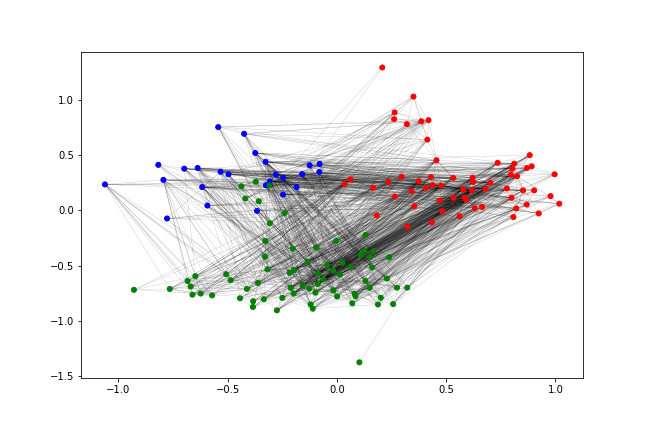}
        \caption{Observed: \textit{Enron} normal operations} 
        \label{fig:enroncontrol_a} 
      \end{subfigure}
      \begin{subfigure}[b]{0.5\linewidth}
        \centering
        \includegraphics[width=\linewidth]{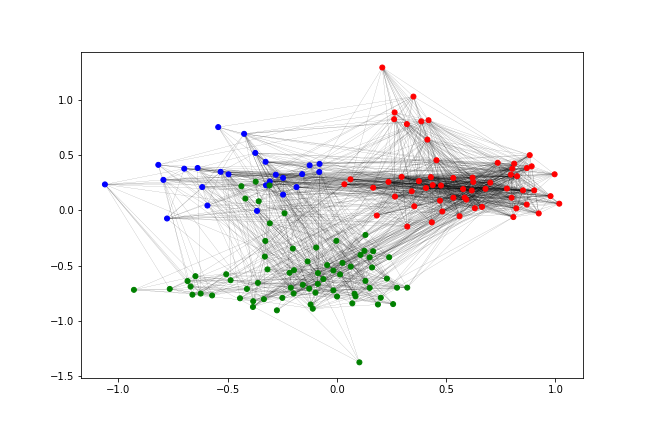} 
        \caption{Controlled: Legal (red) internal surge} 
        \label{fig:enroncontrol_b} 
      \end{subfigure} 
      \begin{subfigure}[b]{0.5\linewidth}
        \centering
        \includegraphics[width=\linewidth]{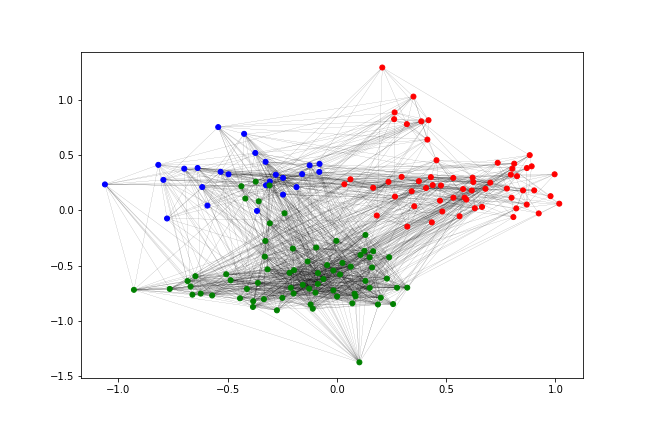} 
        \caption{Controlled: Finance (green) internal surge} 
        \label{fig:enroncontrol_c} 
      \end{subfigure}
      \begin{subfigure}[b]{0.5\linewidth}
        \centering
        \includegraphics[width=\linewidth]{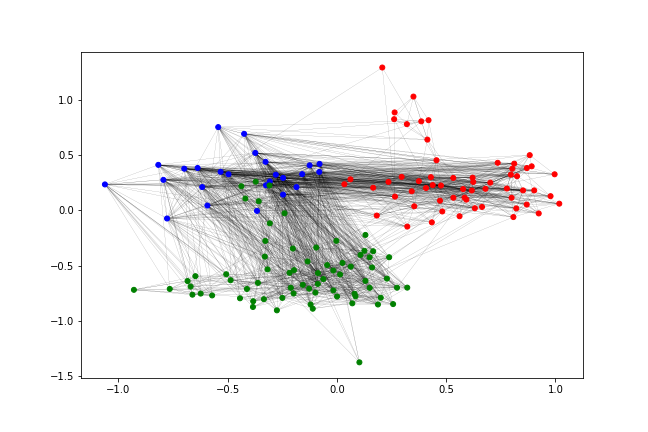} 
        \caption{Controlled: Trading (blue) outgoing surge} 
        \label{fig:enroncontrol_d} 
      \end{subfigure} 
      \end{adjustbox}
    \caption{\textsc{ShadowCast} generated desired graphs of the \textit{Enron} email network. (a) The original graph, (b) the controlled graph of legal (red) team internal surge, (c) the controlled graph of finance (green) team internal surge, and (d) the controlled graph of trading (blue) team outgoing surge.}
    \label{fig:enron_control} 
    \end{figure*}
    \begin{table*}[htp]
    \centering  
    \caption{Performance evaluation of the controlled generation of desired graphs by specifying the Markov model parameters. The \textit{Observed} row shows several of the input graph's important statistical properties, while the rows below contain the values of \textit{absolute differences} between the given original graph and generated graphs. Smaller values point to a closer similarity of generated graphs and the observed graph, thus retaining the original graph's intrinsic structural properties.
    \label{table:customstats}}    
    \resizebox{0.4\linewidth}{!}{%
    \begin{tabular}{c|cccc}
        \textbf{Graph} & 
        \multicolumn{1}{c}{\textbf{\begin{tabular}[c]{@{}c@{}}CLUST\end{tabular}}} &
        \multicolumn{1}{c}{\textbf{\begin{tabular}[c]{@{}c@{}}CPL\end{tabular}}} &
        \multicolumn{1}{c}{\textbf{\begin{tabular}[c]{@{}c@{}}GINI\end{tabular}}} &
        \multicolumn{1}{c}{\textbf{\begin{tabular}[c]{@{}c@{}}MD\end{tabular}}} 
        \\ \hline

        Observed (Figure~\ref{fig:enroncontrol_a})  &\textit{0.03300} &\textit{2.154} &\textit{0.281} &\textit{74.0} \\ \cline{2-5}
        Controlled (Figure~\ref{fig:enroncontrol_b})    &0.00932	&0.079	&0.008	&8.0    \\
        Controlled (Figure~\ref{fig:enroncontrol_c})    &0.00095	&0.104	&0.011	&8.0 \\
        Controlled (Figure~\ref{fig:enroncontrol_d})    &0.00287	&0.084	&0.036	&6.0 \\
    \end{tabular}   }
    \end{table*}

\paragraph{Controlling Generated Graphs}
In addition to recreating graphs that closely match the statistics of the input graphs, we demonstrate our model's ability to generate desired graphs by controlling the parameters of shadows. The shadow allows us a way to gain insight into how graphs are generated and provide controllability. We influence the generative process by constructing shadow walks of preferred distribution using shadow caster $C$. First, we create sequences of node ground-truth labels by specifying the parameters of a transparent and straightforward Markov model: (1) initial probability distribution over $K$ labels $\bm{\pi} = (\pi_1, \pi_2, \dots , \pi_{K})$, where $\pi_i$ is the probability that the Markov chain will start from label $i$, and (2) transition probability matrix $\bm{A} = (a_{11}a_{12}\dots a_{k1}\dots a_{kk})$, where each $a_{ij}$ represents the probability of moving from label $i$ to label $j$. Next, we input the constructed sequences into shadow caster $C$, which returns model-generated shadow walks. Finally, by injecting these designed shadows into our trained generator $G$, we generate desired graphs of different structures.

    \begin{figure*}[htp] 
    \centering 
    \begin{adjustbox}{minipage=\linewidth,scale=1}
        \begin{subfigure}[b]{0.33\linewidth}
        \centering
        \includegraphics[width=.95\linewidth]{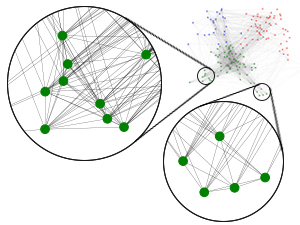}
        \caption{Original graph} 
        \label{fig:subgraph_a} 
        \end{subfigure}
        \begin{subfigure}[b]{0.33\linewidth}
        \centering
        \includegraphics[width=.95\linewidth]{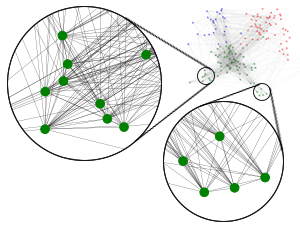}
        \caption{Controlled graph by \textsc{ShadowCast}} 
        \label{fig:subgraph_b} 
        \end{subfigure} 
        \begin{subfigure}[b]{0.33\linewidth}
        \centering
        \includegraphics[width=.95\linewidth]{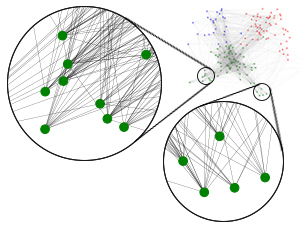} 
        \caption{Random graph construction} 
        \label{fig:subgraph_c} 
        \end{subfigure}
        \end{adjustbox}
    \caption{(a) Subgraph of the original \textit{Enron} email network, (b) the corresponding subset of the graph (Figure~\ref{fig:enroncontrol_c}) generated by \textsc{ShadowCast}, and (c) the respective subset of the random graph construction.}
    \label{fig:controlproperties} 
    \end{figure*}

In Figure~\ref{fig:enron_control}, we show controlled generation examples of the \textit{Enron} email network, where each employee represented by a node belongs to one of three departments (e.g., Legal, Trading, and Finance offices in the organization). Figure~\ref{fig:enroncontrol_a} is an observed instance of interactions between the departments during normal operations. Due to limited observations, network data of some unprecedented, extraordinary situations may be unavailable. To simulate such occurrences, we can set the distribution of the Legal (red), Trading (blue), and Finance (green) departments with parameters ($\bm{\pi}, \bm{A}$) to control the generative process. Distribution configurations $\bm{\pi} = (\pi_1, \pi_2, \pi_3)$ correspond to how likely a sequence of model-generated shadow walks start from a particular department, while the transition probability matrix $\bm{A} = (a_{11}a_{12}\dots a_{31}\dots a_{33})$ determines the probability of moving from one department to another. Various configurations ($\bm{\pi}, \bm{A}$) correspond to different cases such as (Figure~\ref{fig:enroncontrol_b}) internal communication surge in the legal team during court pre-trial period, (Figure~\ref{fig:enroncontrol_c}) internal surge in the finance department during financial accounts reporting period, and (Figure~\ref{fig:enroncontrol_d}) increased outgoing communication between the trading team and the other two departments when purchasing a subsidiary trading firm. Thus, by specifying these parameters, we can control and understand the structure of the generated graphs (see Appendix for the specific parameter settings).

Following the example in Figure~\ref{fig:enroncontrol_c}, one might argue that we could also naively create the effect of an internal email surge by randomly removing the finance (green) inter-department edges and adding random intra-department links. While a simple random graph construction could appear legitimate, it tends to have glaring shortcomings. Specifically, it is not clear if this newly formed graph (1) preserves the intrinsic properties of the original network and (2) has an understandable structure. 
To illustrate the shortcomings, we highlight in Figure~\ref{fig:controlproperties}, circles on the graphs, that the controlled graph (Figure~\ref{fig:subgraph_b}) generated by \textsc{ShadowCast} preserves the original graph's intrinsic properties, whereas the random graph (Figure~\ref{fig:subgraph_c}) generated by the \citet{Erdos:1960} model fails to retain those properties.

In contrast, our approach follows a transparent and straightforward Markov model, providing control over the generated graphs that are modeled based on the original graph. This intuitive approach allows for an increased understanding of the generated graphs. We present a comparison of the graph statistics (\cref{table:customstats}), showing small absolute differences between the given original graph and generated graphs, noting their similarities and retaining the original graph's intrinsic properties.

\section{Conclusion}
\label{conclude}
In this work, we present \textsc{ShadowCast}, a novel controllable graph generative model, which generates designed graphs that are intuitive and understandable. To the best of our knowledge, this method is the first of its kind to specifically address the unique problem of controlling the generative process to produce coherent desired structures of generated graphs. 
Our model demonstrates how it can leverage graph attributes, directed with a transparent Markov model, as a control vector and allow for adjustable parameters to influence the generative process. By introducing controllability in graph generation, a meaningful problem for a better understanding of generated graph data, we hope to encourage further investigation in this line of work and expand on its applications in different areas.


\bibliography{reference}
\bibliographystyle{icml2021}


\end{document}